\documentclass[11pt]{article}

\usepackage[preprint]{acl}

\usepackage{amssymb}
\usepackage{times}
\usepackage{latexsym}
\usepackage{hyperref}
\usepackage{url}
\usepackage{graphicx}
\usepackage{xspace}
\usepackage{algorithm}
\usepackage{algorithmic}
\usepackage{amsmath}
\usepackage{booktabs}
\usepackage{multirow}
\usepackage{array}
\usepackage{pgfplots}
\pgfplotsset{compat=1.18}
\usepackage[T1]{fontenc}

\usepackage[utf8]{inputenc}

\usepackage{microtype}

\usepackage{inconsolata}
\usepackage{listings}
\lstdefinestyle{prompt}{basicstyle=\ttfamily\scriptsize, breaklines=true, breakatwhitespace=true, columns=fullflexible, keepspaces=true, frame=single, framerule=0.3pt, rulecolor=\color{gray!60}, xleftmargin=3pt, xrightmargin=3pt, framesep=3pt, aboveskip=4pt, belowskip=6pt, breakindent=0pt}

\usepackage{graphicx}

\newcommand{\projectname}{\textsc{SENTINEL}\xspace} 
\title{SENTINEL: Weakness-Targeted Task Evolution for Training Multi-Turn Tool-Using LLM Agents}

\author{
 \textbf{Ziyi Wang\textsuperscript{1}},
 \textbf{Yuxuan Lu\textsuperscript{1}},
 \textbf{Yimeng Zhang\textsuperscript{2}},
 \textbf{Yiming Qian\textsuperscript{2}},
 \textbf{Chen Luo\textsuperscript{2}},
 \textbf{Jiri Gesi\textsuperscript{2}},
\\
 \textbf{Hanqing Lu\textsuperscript{2}},
 \textbf{Yisi Sang\textsuperscript{2}},
 \textbf{Huimin Zeng\textsuperscript{3}},
 \textbf{Manling Li\textsuperscript{4}},
 \textbf{Kai-Wei Chang\textsuperscript{5}},
 \textbf{Dakuo Wang\textsuperscript{1}}
\\
 \textsuperscript{1}Northeastern University,
 \textsuperscript{2}Independent Researcher,
 \textsuperscript{3}University of Illinois Urbana-Champaign,
\\
 \textsuperscript{4}Northwestern University,
 \textsuperscript{5}University of California, Los Angeles
}

\begin{document}
\maketitle
\begin{abstract}
Language model agents are increasingly effective at solving realistic tasks through multi-turn tool use, but training reliable tool-using agents remains challenging.
While reinforcement learning provides an on-policy paradigm for improving agents from their own environment interactions, its effectiveness depends heavily on the training tasks.
When training tasks are fixed before training, the task distribution can become increasingly mismatched with the policy's evolving capabilities, causing many rollouts to be spent on uninformative tasks.
We propose \projectname, which tracks the outcome of every training task and keeps generating new tasks that target the current policy's weaknesses, so that the task distribution evolves with the policy.
It makes mastered tasks harder, diagnoses never-solved tasks and builds easier variants around where the agent fails, and turns the failures on partly solved tasks into new tasks.
On $\tau$-Knowledge and $\tau^2$-Bench, \projectname improves the pass rate over GRPO across three domains.
With SFT followed by \projectname, the pass rate of Qwen3.5-27B on $\tau$-Knowledge improves from 25.0\% to 60.8\%.

\end{abstract}
\section{Introduction}

\begin{figure}[t]
    \centering
    \includegraphics[width=\linewidth]{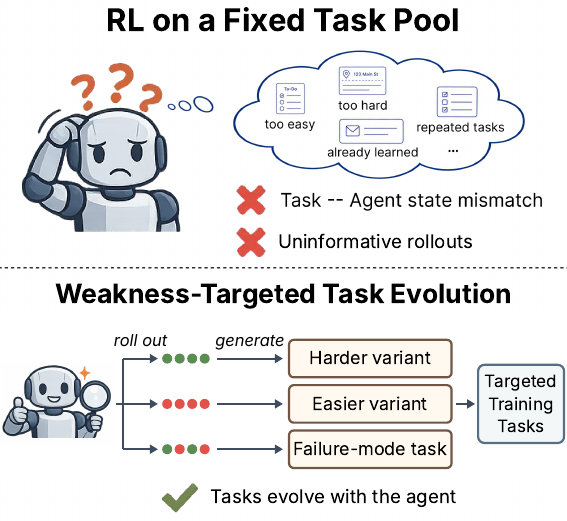}
    \caption{On a fixed task pool, many tasks become too easy or stay too hard for the agent, and rollouts on them carry little signal. \projectname tracks the outcome of each task and generates harder variants of mastered tasks, easier variants of never-solved ones, and new tasks from the failure modes of tasks solved only sometimes. Each row of dots is a group of rollouts on one task: all pass, all fail, or mixed.}
    \label{fig:teaser}
\end{figure}

Large language model (LLM) agents increasingly complete tasks by interacting with tools, users, and environments over many turns, enabling applications in customer service, workflow automation, and software engineering~\citep{yao2022react,yao2024tau,barres2025tau,bandi2026mcp,yang2024swe}.
To train such agents, reinforcement learning (RL) with verifiable outcomes has become a standard approach~\citep{shao2024deepseekmath,guo2025deepseek,qian2025toolrl,feng2025retool,jin2025search,liu2026customeragentovercomingcontextlimitations}: the agent attempts each task several times in an executable environment and is updated toward the attempts that succeed and away from those that fail.

Group-based methods such as GRPO~\citep{shao2024deepseekmath} derive the learning signal from the contrast among attempts: when all attempts at a task succeed or all fail, there is no contrast between success and failure to learn from.
With a fixed pool of training tasks, the signal is therefore missing at both ends of difficulty: a task the agent has mastered stops contributing because every attempt succeeds, and a task the agent never solves contributes nothing because every attempt fails.
Dynamic sampling~\citep{yu2026dapo} and task-selection curricula~\citep{jiang2021prioritized} only choose among the existing tasks, so they cannot create signal that the pool does not provide.
Some recent methods generate new training tasks, either through self-play, where the model proposes tasks for itself~\citep{zhao2026absolute,huang2025r,xia2025agent0,zhou2026self}, or from the agent's own failed rollouts~\citep{qi2025webrl,liang2025sws,yang2026coevolve}.
However, none of these methods acts on both ends of difficulty: a mastered task shows a skill that a harder variant can test further, and a never-solved task points to a weakness, such as misapplying an eligibility threshold, that an easier variant can isolate.
Moreover, generating a valid task itself is hard in multi-turn, stateful environments, because each task must specify the environment state, the policy, the user's behavior, and a verifiable outcome consistently, or the task becomes unsolvable or its reward wrong.

We propose \projectname, which keeps RL training at the agent's frontier by evolving the training tasks to target the agent's current capability (Figure~\ref{fig:teaser}).
\projectname enables valid task generation in policy-rich, multi-turn tool-calling environments.
During training, \projectname tracks the outcome of every task and sorts it into one of three cases: mastered, never solved, or solved only sometimes.
For a mastered task, \projectname generates a harder variant that tests the same skill under more demanding conditions.
For a never-solved task, it first diagnoses why the agent fails and then generates an easier variant that targets the same weakness but is within the agent's reach.
Once the agent masters the easier variant, harder variants lead it back toward the original task.
For a task solved only sometimes, its failed rollouts are grouped into failure modes, each of which seeds new tasks.
In addition, exploration proposes tasks for situations that no rollout has covered yet.

We conduct experiments on $\tau$-Knowledge~\citep{shi2026tau}, a banking environment whose policies span a large knowledge base, and on the retail and airline domains of $\tau^2$-Bench~\citep{barres2025tau}, where the agent handles orders and flight reservations under a written domain policy.
\projectname outperforms GRPO in every domain, by 5.0 points in banking with Qwen3.5-27B and 2.5 with Qwen3.5-9B, 3.3 in retail, and 5.0 in airline.
Throughout training, it keeps more rollout groups informative.
Our contributions are as follows:
(1) We propose \projectname, a weakness-targeted task evolution method that generates valid tasks in multi-turn, stateful environments; by matching the agent's current capability, these tasks provide more learning signal than a fixed task pool.
(2) We introduce a policy-grounded task construction pipeline that extracts a short list of the policy rules deciding a situation, derives the task from these rules, and builds a new environment state in which they apply. This enables generating both RL tasks from the agent's weaknesses and SFT data from seed situations.
(3) Empirical results on three domains show that \projectname consistently outperforms GRPO, and our analyses show that the generated tasks provide more learning signal than the original tasks.

\section{Related Work}

\subsection{Agentic Reinforcement Learning}
Agentic reinforcement learning (RL) trains a language model as a policy that interacts with an environment over multiple steps~\citep{zhou2024archer,wang2025ragen,zhang2025landscape}.
At each step, the policy takes an action and receives feedback from the environment, which informs its next action; the whole trajectory is then optimized with task-level rewards on the final outcome or process-level rewards on intermediate steps~\citep{guo2025deepseek,qian2025toolrl,wang2025spa}.
Agentic RL has been applied to tool-integrated reasoning~\citep{qian2025toolrl,li2025torl,feng2025retool}, search~\citep{jin2025search}, web and device control~\citep{qi2025webrl,bai2024digirl}, software engineering~\citep{golubev2025training,da2025agent}, online shopping~\citep{zhang2025shop,wang2025customer,liu2026customeragentovercomingcontextlimitations}, and multi-turn tool use with simulated users~\citep{gao2026self}.
Many of these works use group-based algorithms such as GRPO~\citep{shao2024deepseekmath}, which sample several attempts at each task and learn from the differences in their rewards; a task whose attempts all receive the same reward contributes no policy gradient.
DAPO~\citep{yu2026dapo} filters out such groups and keeps sampling until the batch is full of informative ones, and other methods select tasks by their difficulty or learning progress~\citep{bae2025online,foster2025learning,graves2017automated,jiang2021prioritized}.
These methods choose among existing tasks and do not add new ones once the pool runs out of informative tasks; \projectname generates new tasks instead.

\subsection{Task Synthesis for Tool Use}
Since large-scale human-annotated tool-use data are costly to collect, synthetic data generation has become a common strategy for training tool-using agents.
ToolAlpaca generates tool-calling demonstrations from tool descriptions~\citep{tang2023toolalpaca}, APIGen-MT synthesizes multi-turn tool-use data in simulated environments~\citep{prabhakar2025apigen}, AReaL-SEA synthesizes multi-turn tool-use dialogues together with an executable checker for each dialogue~\citep{gao2026self}, and recent work scales synthesis to MCP-based environments~\citep{wang2025mcp,xu2025toucan}.
A key challenge is verifiability: a generated task should correspond to a feasible tool-call path with correct intermediate states and outcomes.
Several methods therefore start from executable trajectories rather than asking a model to write both a task and its solution: Trajectory2Task builds valid trajectories first and converts them into user tasks~\citep{wang2026trajectory2tasktrainingrobusttoolcalling}, and Firefly explores real APIs and derives tasks backward from the observed outcomes~\citep{lu2026firefly}.
Other methods ground tasks in the environment's rules and state: IntellAgent generates policy-driven scenarios for evaluating conversational agents~\citep{levi2025intellagent}, GAIS builds environments from real MCP servers and plans tasks over them, enforcing logical dependencies between tool calls and injecting adversarial policies~\citep{shi2026gais}, EdgeGen generates database-grounded edge cases designed to violate the compliance rules in an agent's specification~\citep{abichandani2026edgegen}, and KARL synthesizes grounded search tasks over enterprise documents with an agentic pipeline~\citep{chang2026karl}.
These methods build tasks for training or evaluating tool-using agents, but they do not build them from the weaknesses that the agent shows during training.
\projectname instead generates tasks during RL from the current policy's weaknesses, and its construction builds stateful tasks on rules it discovers in the domain's policy documents.

\subsection{Online Task Generation}
Another line of work generates new tasks while the agent trains, so that the tasks keep up with the agent's ability.
Self-play methods let the model propose tasks for itself: in reasoning, at a difficulty it can learn from~\citep{zhao2026absolute,huang2025r,xia2025agent0}, and in tool use, with code that verifies each task~\citep{zhou2026self}.
Failure-driven methods instead generate new tasks from the agent's weaknesses: from its failed rollouts for web agents~\citep{qi2025webrl} and for general LLM agents~\citep{yang2026coevolve}, from the concepts of problems it fails in math reasoning~\citep{liang2025sws}, and from its success rate on each skill for embodied agents~\citep{zala2024envgen}.
\projectname acts on the outcome of each task: besides new tasks that target failure modes, it builds harder variants of mastered tasks and, for a task the agent never solves, an easier variant around the step where the agent fails; all of them are stateful, natural-language tasks.

\section{Method}
\label{sec:method}

\begin{figure*}[t]
    \centering
    \includegraphics[width=\textwidth]{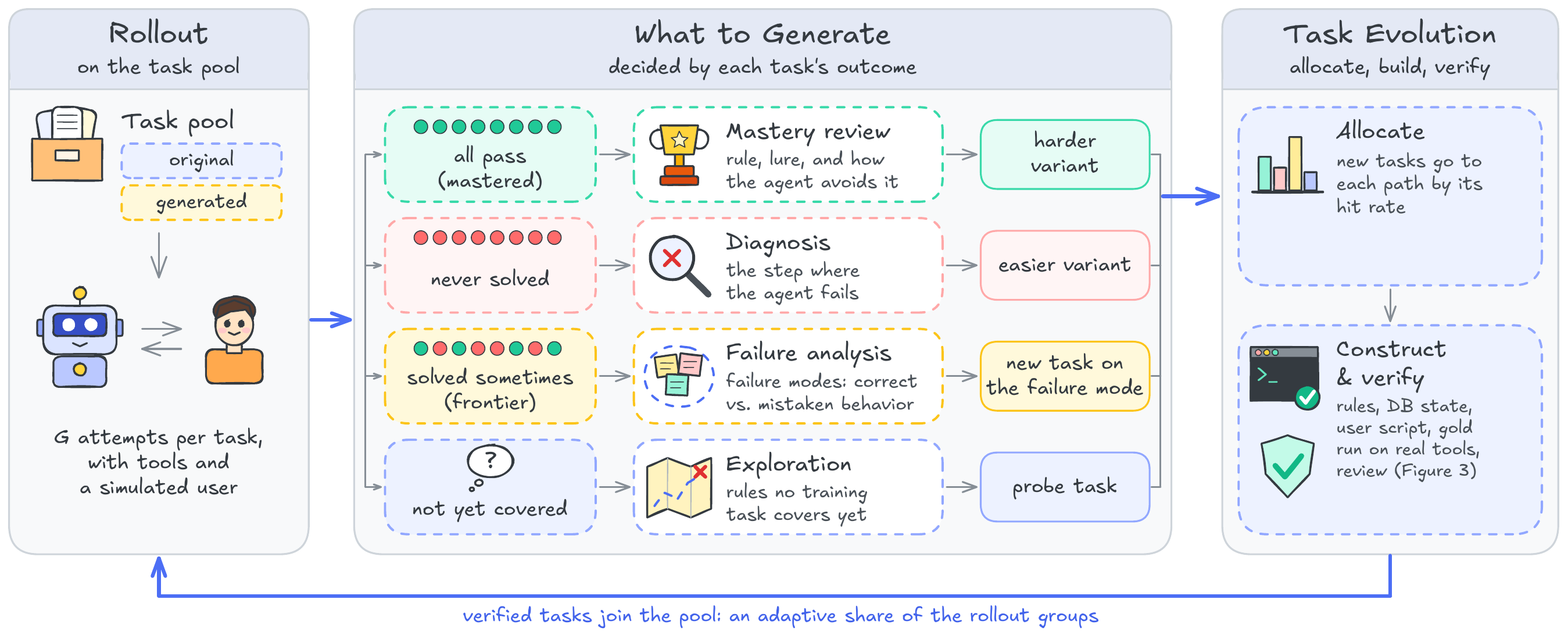}
    \caption{Overview of \projectname. Every rollout group updates the outcome of its task, which decides the task's generation path: a mastered task gets a harder variant, a never-solved task is diagnosed and gets a decomposed easier variant, failures on frontier tasks are grouped into failure modes that seed new tasks, and exploration probes rules that no training task covers yet. Each round allocates its generation jobs among the paths by how often their earlier tasks produced mixed groups. Every task is constructed and verified in the environment (Figure~\ref{fig:construction}) before it enters the pool, which serves generated tasks at an adaptive share of the rollout groups.}
    \label{fig:overview}
\end{figure*}

Figure~\ref{fig:overview} shows the overview of \projectname. \projectname tracks the outcome of every training task and keeps generating new tasks that target the current policy's weaknesses, so the task pool becomes a curriculum that adapts to the policy as it trains. We first define tasks (\S\ref{sec:setup}), then describe how the outcome of each task decides what to generate (\S\ref{sec:paths}), and how new tasks are constructed and verified (\S\ref{sec:construction}). 

\subsection{Problem Setup}
\label{sec:setup}

A task $x = (E, u, c)$ consists of an environment $E$, a user intention $u$, and a checker $c$. The environment provides tools, a domain policy, and a database set to the task's initial state. The user intention states what the user wants done and instructs a simulated user how to behave. The agent interacts with this simulated user over multiple turns, following the policy and calling tools to look up information and modify the database. When the conversation ends, $c$ checks the final database and the information given to the user, and outputs a binary success $y \in \{0, 1\}$.

\subsection{Weakness-Targeted Task Generation}
\label{sec:paths}

\paragraph{Outcome states.}
During RL training, after each rollout group of a task, the task is \emph{mastered} if every attempt in the group succeeds, \emph{never solved} if it has received at least two full groups without a single success, and at the \emph{frontier} otherwise. \projectname generates a different kind of new task for each state, and exploration adds a fourth path that does not depend on any task. Appendix~\ref{app:generation} gives the triggers and limits of each path.

\paragraph{Analyses.}
Each path has its own analysis by an LLM: a mastery review for mastered tasks, a diagnosis for never-solved tasks, a failure analysis for frontier tasks, and an exploration plan (Appendix~\ref{app:prompts} lists the prompts). The mastery review, the diagnosis, and the failure analysis see the task (user intention, initial state, and expected actions), the agent's trajectory (dialogue, tool calls, and tool results), its outcome, and the domain policy. The exploration plan sees no trajectory; it sees the current training status of each weakness found so far and the domain rules that no training task uses yet. Every analysis describes what it finds as a \emph{potential weakness} of the agent: the situation, the correct behavior, the mistaken behavior, and the rule that decides between them. The following paragraphs describe each path.

\paragraph{Mastered: harder variants.}
The first time a task is mastered, a mastery review names the weakness that the task trains and how the agent now avoids the mistake, and specifies how to make the task harder on the same rule. A harder variant keeps the rule and makes the mistake easier to make, with a stronger lure in the user's script or in the database records. For example, after the agent masters computing a savings rate from its stacked bonuses, the harder variant's records let the balance drop below \$10,000, the threshold of the higher base rate, for half of the month, and the customer insists on the full rate. The agent thus practices the same rule in harder situations.

\paragraph{Never solved: decomposed easier variants.}
A never-solved task is first diagnosed from the agent's failed attempts: the diagnosis finds the step at which the agent fails and the rule that it breaks there. The easier variant trains only this step: it removes the rest of the task, such as the other requests that the agent already handles, and keeps the rule that decides this step. For example, in a banking task, a customer whose debit card was locked after their own failed PIN attempts asks to unlock it; the agent must refuse, close the card, and order a replacement, but it wrongly says that expedited shipping and a premium card design are free. The easier variant asks only for the replacement card. Once the agent masters an easier variant, the mastered path gives that variant a harder variant, so the difficulty climbs back toward the original task. If a variant misses its goal, that is, a harder variant is still mastered or an easier one is still never solved, the next variant must change more than surface details.

\begin{figure*}[t]
    \centering
    \includegraphics[width=\textwidth]{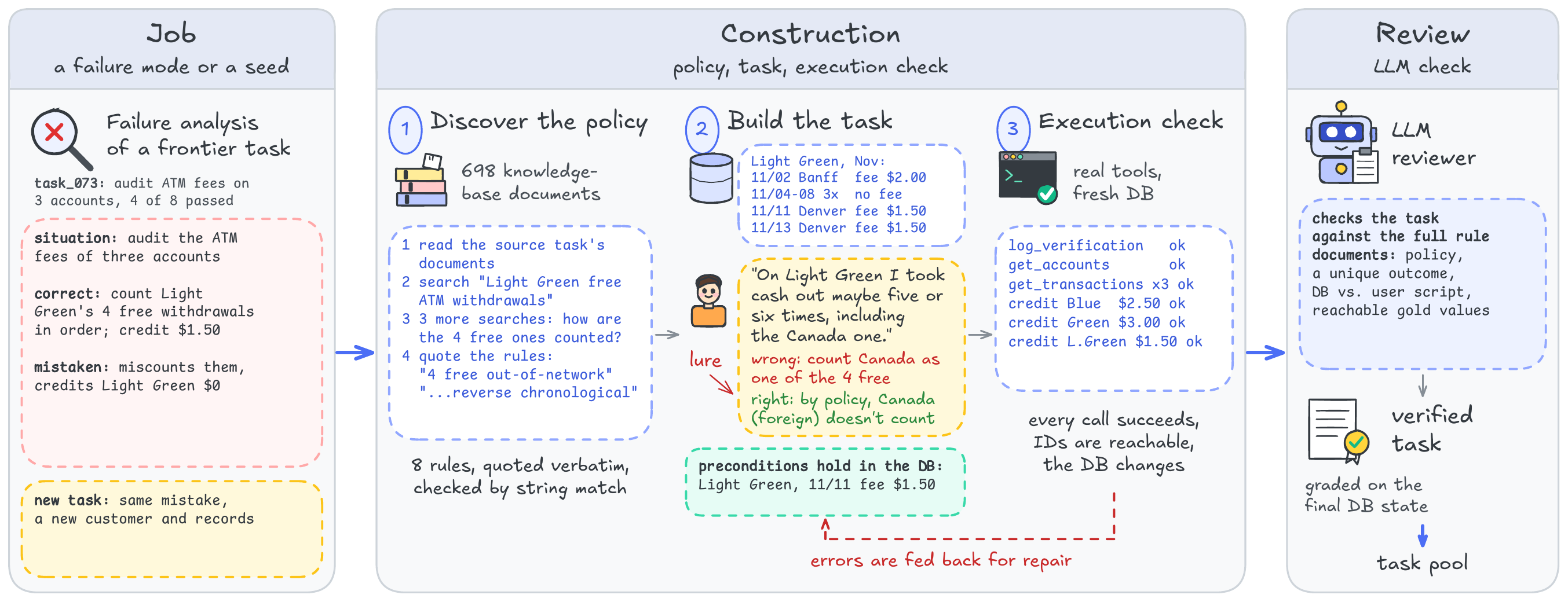}
    \caption{Policy-grounded task construction, shown on a task built from a failure mode of a frontier banking training task. The job carries the failure mode; for SFT data, a seed situation takes its place. The constructor discovers the policy by searching the knowledge base step by step and quotes the rules verbatim, writes the database state and a user script whose lure invites the mistake, and runs the gold actions on the environment's own tools; failed checks are fed back for repair. An LLM reviewer then checks the task against the full rule documents before it joins the pool.}
    \label{fig:construction}
\end{figure*}

\paragraph{Frontier: failure modes.}
A frontier task is solved only sometimes, so its failed attempts show mistakes that the agent is close to overcoming. The failure analysis groups the failures of each round into failure modes, the weaknesses that these failures show (Table~\ref{tab:modes}). It discovers the failure modes from the trajectories instead of choosing them from a fixed taxonomy, and it sees the failure modes of recent rounds, so that a mistake that recurs is assigned to the same failure mode. For example, in one failure mode, a customer adds up their cash back wrongly and insists that the bank underpaid them, and the agent gives in and files a rewards dispute that the records do not support. \projectname then builds new tasks for each failure mode: the policy stays the same, and each task gives a new customer with new records a situation in which the same mistake can occur. For this failure mode, a generated task could be a new customer who keeps pushing for a different infeasible request, which the agent must refuse. Appendix~\ref{app:example} shows an example.

\paragraph{Exploration.}
Failures reveal weaknesses only in situations that the training tasks already cover. Exploration therefore looks at the domain rules that no training task uses yet and at the weaknesses found so far with their training status. From these, it proposes potential weaknesses in situations not covered so far, each with the documents that govern it.

\paragraph{Task allocation.}
A generated task is useful for training only if its groups are mixed, i.e., the agent solves it sometimes but not always. Some paths produce such tasks more often than others, and which paths do so changes as the agent improves. \projectname therefore treats the choice of path as a non-stationary multi-armed bandit~\citep{graves2017automated,chen2025self}, in which each path is an arm and a task with a mixed group is a reward. Generation runs in rounds, one after every $N$ rollouts, and each round builds $J$ new tasks. A generated task becomes a \emph{hit} once it produces a mixed group, and a \emph{miss} once it produces an all-pass group or two groups without a mix. Let $h_k$ and $m_k$ be the numbers of hits and misses among the last $W$ tasks of path $k$ that have become a hit or a miss. The path's \emph{hit rate} $r_k$ and the probability $P(k)$ that a new task goes to path $k$ are
\begin{align}
r_k &= \frac{h_k + \alpha}{h_k + m_k + 2\alpha}, \\
P(k) &= (1 - \epsilon)\,\frac{r_k}{\sum_{j=1}^{K} r_j} + \epsilon \cdot \frac{1}{K},
\end{align}
where $K$ is the number of paths that can build a task in this round. The prior $\alpha$ handles the cold start: a path with few hits and misses starts with a hit rate near one half. The uniform term with weight $\epsilon$ keeps trying a path whose recent tasks missed, so that its hit rate can recover. Figure~\ref{fig:alloc} shows how the allocation shifts during training.

\subsection{Policy-Grounded Task Construction}
\label{sec:construction}

An LLM constructor builds each new task according to the potential weakness and analysis from \S\ref{sec:paths}. It follows four steps (Figure~\ref{fig:construction}): it finds the policy rules that decide the situation, writes the task and its gold actions from these rules, executes the gold actions on the environment, and passes the task to an independent reviewer. The expected outcome thus comes from the rules rather than from a model attempting the task. Appendix~\ref{app:construction} gives more details.

\paragraph{Policy discovery.}
The policy is often complex and spread over many documents, so the constructor first extracts the part of it that the task relates to. It starts from the documents of the source task, or from those that the weakness names. When these do not settle the situation, it looks further one step at a time: each step either searches the policy with a query, which returns the few best-matching documents in full, or opens one document by its identifier. It stops once the documents it has seen settle the situation, or after a fixed number of steps; when the policy is a single document, it simply reads all of it. It returns a list of rules in two groups: the rules that decide the outcome, and the rules needed around them, such as identity verification, eligibility, fees, and which tool performs the action. Each rule states a requirement and quotes its supporting text from the policy, so that the next steps work from this list and the documents it cites rather than from the whole policy.

\paragraph{Build the task.}
The constructor then builds the task around these rules. It sets up an environment in which the deciding rules apply, by writing a new customer and their records, either into an empty database or on top of a database of synthetic records, so the task shares no customer or record with the original tasks. It writes the user intention: what the customer asks for and how they behave in the conversation. To invite the mistaken behavior, it adds a lure, either to the user's instructions (the user insists, changes their mind, or reveals the deciding detail only when asked) or to the records (a value just past a threshold, or look-alike accounts). Finally, it writes the gold actions following the listed rules: the tool calls that a correct agent makes on these records.

\paragraph{Execution check.}
Gold actions written from the rules in one pass can still miss details or not match the arguments that the tools expect, so the constructor first runs them on a fresh copy of the environment state with the environment's own tools. Every call must succeed, every identifier that the agent has to supply must be obtainable from the records or the conversation, and every piece of information that the agent must tell the user must appear in a tool output. Appendix~\ref{app:construction} gives the further checks of each benchmark. Any violation is fed back to the constructor for repair, up to a fixed number of attempts.

\paragraph{Review.}
Finally, an independent LLM reviewer checks the task against the full policy: the correct outcome must be unique and allowed by the rules, the user's instructions must give everything the user needs to say and agree with the records and the gold, and solving the task must require the correct behavior of the weakness it targets. Any blocking issue rejects the task. An accepted task then joins the training pool, which serves it at an adaptive share of the rollout groups (Appendix~\ref{app:pool}).

\paragraph{Offline data construction.}
The same four steps also run offline, without RL rollouts: they take as input a seed situation, a customer goal that we write from the policy, instead of a potential weakness. We use them to build the SFT data (Appendices~\ref{app:construction} and~\ref{app:sft}).

\section{Experiments}
\label{sec:experiments}

\subsection{Experimental Setup}
\label{sec:exp-setup}

\paragraph{Benchmarks.}
We evaluate on $\tau$-Knowledge~\citep{shi2026tau}, which we call banking, and on the retail and airline domains of $\tau^2$-Bench~\citep{barres2025tau}. In banking, the agent serves bank customers by searching a knowledge base of 698 policy documents; in retail and airline, it follows a written domain policy. Appendix~\ref{app:benchmarks} gives the benchmark details, including the train/test splits and the grading defects that we fix in banking. Test tasks are never used for task generation.

\paragraph{Training.}
Each domain starts from a supervised fine-tuned checkpoint: Qwen3.5-27B and Qwen3.5-9B~\citep{qwen2026qwen35} for banking, both fine-tuned on our SFT data (Appendix~\ref{app:sft}), and Qwen3-4B-Thinking-2507~\citep{yang2025qwen3} for retail and airline, fine-tuned on the Trajectory2Task dataset~\citep{wang2026trajectory2tasktrainingrobusttoolcalling}. We train with GRPO, using Dr.GRPO advantages~\citep{liu2025understanding}, $G = 8$ attempts per task, dynamic sampling~\citep{yu2026dapo}, and the reward described in Appendix~\ref{app:reward}. Kimi-K3~\citep{kimi2026k3} simulates the user in training and evaluation, and also performs \projectname's analyses and task construction. \projectname starts a generation round every $N$ rollouts and builds $J$ new tasks per round: $N = 128$ and $J = 4$ in banking with the 27B model, $N = 128$ and $J = 8$ with the 9B model, and $N = 64$ and $J = 8$ in retail and airline. It uses a hit-rate window of $W = 30$ tasks, a prior of $\alpha = 2$, and $\epsilon = 0.1$. The GRPO baseline uses the same settings on the original training tasks only. Appendices~\ref{app:generation}, \ref{app:sft}, and \ref{app:prompts} give the generation details, the SFT data, and the prompts.

\paragraph{Evaluation.}
We run each checkpoint three times on the test tasks, and four times in airline, since airline has fewer test tasks. A task is graded on the final database and the information given to the user, rather than by an LLM judge, which can be biased or miscalibrated~\citep{zheng2023judging,liu2026calibrating}. We report the pass rate averaged over the trials. For each trained model, Tables~\ref{tab:main} and~\ref{tab:ablation} report the checkpoint with the highest pass rate.

\subsection{Main Results}
\label{sec:main-results}

\begin{table}[t]
    \centering
    \small
    \begin{tabular}{lcccc}
        \toprule
         & \multicolumn{2}{c}{Banking} & Retail & Airline \\
        \cmidrule(lr){2-3}
        Method & 27B & 9B & 4B & 4B \\
        \midrule
        Kimi-K3 & \multicolumn{2}{c}{53.3} & 82.5 & 75.0 \\
        \midrule
        Base model & 25.0 & 9.2 & 73.3 & 56.2 \\
        SFT start & 48.3 & 39.2 & 75.8 & 68.8 \\
        GRPO & 55.8 & 45.0 & 85.0 & 76.2 \\
        \projectname & \textbf{60.8} & \textbf{47.5} & \textbf{88.3} & \textbf{81.2} \\
        \bottomrule
    \end{tabular}
    \caption{Test pass rate (\%). Banking is from $\tau$-Knowledge, with Qwen3.5 at 27B and 9B; retail and airline are from $\tau^2$-Bench, with Qwen3-4B. Bold marks the best trained model.}
    \label{tab:main}
\end{table}

\begin{figure*}[t]
    \centering
    \includegraphics[width=\textwidth]{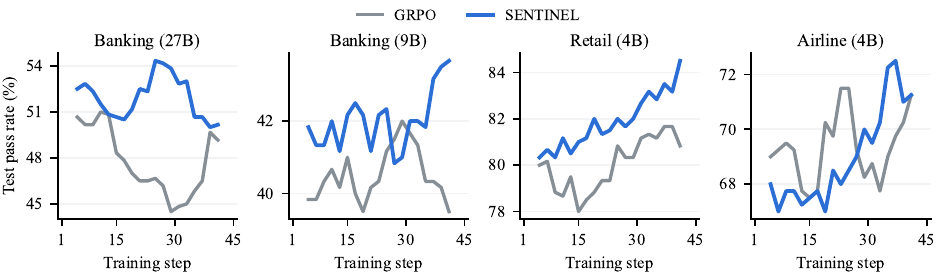}
    \caption{Test pass rate during training, as a moving average over five consecutive checkpoints, which are saved every two steps.}
    \label{fig:curve}
\end{figure*}

Table~\ref{tab:main} compares the base model, the SFT starting point, GRPO trained on the original training tasks, and \projectname. SFT alone already improves the base model substantially: by 23.3 and 30.0 points in banking with the 27B and 9B models and by 12.6 points in airline, though only by 2.5 in retail. From this stronger start, GRPO adds 5.8 to 9.2 points, but it trains only on the fixed task pool, where a growing share of rollout groups carries no signal as the agent improves (\S\ref{sec:signal}). \projectname pushes further in every setting, by 5.0 points over GRPO in banking with the 27B model and 2.5 with the 9B model, 3.3 in retail, and 5.0 in airline. It also surpasses Kimi-K3 in every setting except banking with the 9B model, although Kimi-K3 constructs and reviews its generated tasks. This suggests that training on verified, weakness-targeted tasks can lift an agent beyond the model that builds them: the constructor has the deciding rules in hand and verifies the gold actions by execution, whereas the agent has to find the rules itself in a conversation with a user who may push back. Figure~\ref{fig:curve} shows the test pass rate during training: \projectname reaches a higher pass rate than GRPO in every setting.

\subsection{Where the Signal Comes From}
\label{sec:signal}

\begin{table}[t]
    \centering
    \small
    \setlength{\tabcolsep}{3.5pt}
    \begin{tabular}{llccc}
        \toprule
        & Groups & All pass & Mixed & All fail \\
        \midrule
        \multirow{4}{*}{Banking} & GRPO & 20.1 & 57.7 & 22.2 \\
        & \projectname & 17.4 & \textbf{61.2} & 21.5 \\
        & \quad original tasks & 21.9 & 59.1 & 19.0 \\
        & \quad generated tasks & 11.2 & 63.9 & 24.9 \\
        \midrule
        \multirow{4}{*}{Retail} & GRPO & 41.1 & 56.6 & 2.4 \\
        & \projectname & 33.1 & \textbf{64.1} & 2.8 \\
        & \quad original tasks & 37.6 & 60.2 & 2.1 \\
        & \quad generated tasks & 17.5 & 77.5 & 5.0 \\
        \midrule
        \multirow{4}{*}{Airline} & GRPO & 45.3 & 43.0 & 11.8 \\
        & \projectname & 32.3 & \textbf{56.7} & 11.0 \\
        & \quad original tasks & 43.5 & 44.8 & 11.7 \\
        & \quad generated tasks & 16.7 & 73.3 & 10.0 \\
        \bottomrule
    \end{tabular}
    \caption{Outcomes of all complete rollout groups.}
    \label{tab:signal}
\end{table}

\paragraph{More groups are mixed.}
Table~\ref{tab:signal} counts the outcomes of all complete rollout groups. Only a mixed group contrasts success with failure; in an all-fail group, the rewards differ only by partial credit and penalties (Appendix~\ref{app:reward}), which give a much weaker signal. \projectname has more mixed groups than GRPO in every domain: 61.2\% vs.\ 57.7\% in banking, 64.1\% vs.\ 56.6\% in retail, and 56.7\% vs.\ 43.0\% in airline. The gain comes from the generated tasks, which are mixed far more often than the original ones (63.9\% vs.\ 59.1\% in banking, 77.5\% vs.\ 60.2\% in retail, and 73.3\% vs.\ 44.8\% in airline) and rarely end up all-pass. Under GRPO, the share of all-pass and all-fail groups grows from the first half of training to the second in banking and retail as the agent masters more tasks; in banking, for example, all-pass groups rise from 18.0\% to 22.3\%. In airline, the share stays near 57\% in both halves, mostly because 45.3\% of all groups pass every attempt. \projectname keeps this share lower than GRPO in both halves of training in every domain.

\begin{table}[t]
    \centering
    \small
    \setlength{\tabcolsep}{2.6pt}
    \begin{tabular}{lcccccc}
        \toprule
        & & & & & \multicolumn{2}{c}{First group} \\
        \cmidrule(lr){6-7}
        Path & Jobs & Acc. & Trained & Mixed & All pass & All fail \\
        \midrule
        Frontier & 54 & 50 & 36 & 28 & 2 & 8 \\
        Easier & 18 & 15 & 14 & 8 & 2 & 6 \\
        Harder & 35 & 26 & 22 & 11 & 7 & 4 \\
        Explore & 7 & 6 & 6 & 4 & 2 & 0 \\
        \bottomrule
    \end{tabular}
    \caption{Task generation by path in banking, steps 1--45: generation jobs and accepted tasks; generated tasks trained, those with at least one mixed group, and the outcome of their first group.}
    \label{tab:paths}
\end{table}

\begin{figure*}[t]
    \centering
    \includegraphics[width=\textwidth]{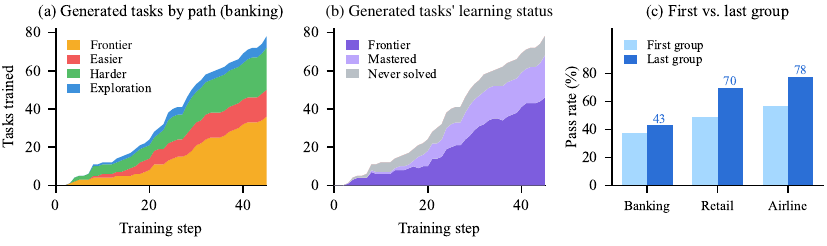}
    \caption{Generated tasks during training. (a) Number of generated tasks trained so far in banking, by the path that produced them. (b) The learning status of the generated tasks over training: at the frontier, mastered, or never solved. (c) For each generated task trained at least twice, the pass rate in its first group, when it was first trained, and in its latest group.}
    \label{fig:analysis}
\end{figure*}

\paragraph{The generated tasks stay informative and are learned.}
Figure~\ref{fig:analysis} follows the generated tasks through training, with the 27B model in banking for Figure~\ref{fig:analysis}a and b. By step 45, 78 generated tasks have been trained, most of them from failure modes (Figure~\ref{fig:analysis}a). As training goes on, more of these tasks are mastered: by step 45, 46 of them are at the frontier, 22 have been mastered, and 10 have never been solved (Figure~\ref{fig:analysis}b). The agent also learns the generated tasks. For every generated task trained at least twice, we compare its pass rate in its first group, when it was first trained, with its pass rate in its latest group (Figure~\ref{fig:analysis}c): the pass rate rises from 38\% to 43\% in banking, from 49\% to 70\% in retail, and from 57\% to 78\% in airline.

\paragraph{The generation paths differ in training signal.}
Table~\ref{tab:paths} breaks generation down by path. \projectname ran 114 generation jobs and accepted 97 tasks (85\%). Tasks from the frontier path are the most productive: 28 of the 36 that were trained produced at least one mixed group, and 74 of their 98 groups were mixed. Easier variants are often still too hard at first (6 of 14 fail their whole first group), which is why a variant that fails its first group is decomposed further right away. Harder variants are sometimes not hard enough (7 of 22 pass their whole first group). The task allocation follows these differences and gives the frontier path the most jobs (Figure~\ref{fig:alloc} in Appendix~\ref{app:analyses}).

\begin{table}[t]
    \centering
    \small
    \begin{tabular}{>{\raggedright\arraybackslash}p{0.33\linewidth}>{\raggedright\arraybackslash}p{0.55\linewidth}}
        \toprule
        Failure mode & Mistaken behavior \\
        \midrule
        Best rate combination (14) & The customer anchors on a headline rate; the agent stops at a worse combination of accounts. \\
        Incomplete multi-item audit (14) & Asked to audit every transaction on four cards, the agent disputes only some of the errors. \\
        Non-stacking bonuses (11) & The agent adds a relationship bonus that does not stack with the account's rate. \\
        Stolen wallet (11) & The agent closes the cards directly instead of freezing them first. \\
        Unwarranted rewards dispute (11) & The customer miscomputes cash back and pushes back; the agent gives in and files a dispute. \\
        Ineligible account opening (11) & The customer pushes for an account they are not eligible for; the agent opens it anyway. \\
        \bottomrule
    \end{tabular}
    \caption{The most frequent of the 47 failure modes found in banking in the first 29 rounds, with the number of failures assigned to each.}
    \label{tab:modes}
\end{table}

\paragraph{Failure modes are specific.}
The failure analysis for the frontier tasks during training finds 47 distinct failure modes (Figure~\ref{fig:modes-discovery} in Appendix~\ref{app:analyses}). Table~\ref{tab:modes} lists the most frequent ones. They name a concrete situation and a concrete mistake, such as acting on only part of a multi-item request, giving in to a customer's pushback, or opening an account that the customer is not eligible for.

\subsection{Ablations}
\label{sec:ablations}

\begin{table}[t]
    \centering
    \small
    \begin{tabular}{lc}
        \toprule
        Method & Banking \\
        \midrule
        GRPO & 55.8 \\
        \projectname & \textbf{60.8} \\
        \quad w/o diagnosis & 55.8 \\
        \quad pre-generated tasks & 56.7 \\
        \quad static curriculum & 56.7 \\
        \quad frontier path only & 60.0 \\
        Offline synthetic tasks only (100) & 57.5 \\
        Offline synthetic tasks only (486) & 58.3 \\
        \bottomrule
    \end{tabular}
    \caption{Banking test pass rate.}
    \label{tab:ablation}
\end{table}

Table~\ref{tab:ablation} compares \projectname with several variants. \emph{W/o diagnosis} builds each new task with only a generic instruction (a new task in the same situation, or an easier or a harder variant), without the agent's trajectory or any analysis. \emph{Pre-generated tasks} runs the generation once before training, from the rollouts of the starting policy on the original tasks, and keeps the resulting tasks fixed. \emph{Static curriculum} reuses all the tasks generated in an earlier \projectname run, together with the original tasks, without generation during training. \emph{Frontier path only} keeps only the frontier path, without easier or harder variants or exploration. \emph{Offline synthetic tasks only} replaces the original tasks with tasks that our task construction pipeline builds offline from seed situations (\S\ref{sec:construction}); unlike pre-generated tasks, they are not built from any rollout of the agent. We first use a random 100 of them, about as many as \projectname trains by step 45 (78), and then all 486 to see whether more tasks help. These 486 tasks come from the first two of the five seed batches in Table~\ref{tab:sft}, so they are a subset of the 932 offline tasks from which we build the SFT data (Appendix~\ref{app:sft}).

Without the diagnosis, the constructor never sees how the agent failed, so its tasks miss the agent's actual mistakes: after step 15, only 49--51\% of the rollout groups on these tasks are mixed, against 57--69\% on \projectname's generated tasks, and the test pass rate falls to that of GRPO. Pre-generated tasks target the weaknesses of the starting policy rather than those of the policy being trained. As the agent improves, it masters the easier pre-generated tasks, which then leave the pool, and no new task replaces them: in steps 31--45, 49.3\% of the groups on pre-generated tasks fail every attempt, against 23.3\% on \projectname's generated tasks. The static curriculum serves, from the first step on, tasks that an earlier run generated at every stage of its training. Many of these reused tasks do not fit the current policy: 42.7\% of the groups on them fail every attempt, dynamic sampling drops 32.1\% of all groups against 18.0\% for \projectname, and the test pass rate stays 4.1 points lower.

The frontier path alone comes within 0.8 points of the full method. Its generated tasks are mixed about as often as \projectname's (63.4\% vs.\ 63.9\% of the groups) and rarely become too easy (4.5\% all-pass), so the tasks from the frontier path carry most of the signal, as \S\ref{sec:signal} also shows. Without the easier path, however, more groups on the frontier-path-only tasks fail every attempt (32.2\% vs.\ 24.9\%), because no easier variant is built for the tasks that the agent cannot solve.

Offline synthetic tasks are not aimed at the agent's failures, and the starting policy already solves many of them: 54.7\% and 51.3\% of the groups on the 486 and the 100 offline synthetic tasks pass every attempt, and dynamic sampling drops over half of all groups. Training on offline synthetic tasks still improves over GRPO, but adding more offline synthetic tasks helps little: the run with 486 offline synthetic tasks reaches 58.3, only 0.8 points above the run with 100, while \projectname, which trains on 78 generated tasks, reaches 60.8. What the tasks target matters more than how many there are.

\section{Conclusion}
\label{sec:conclusion}
We present \projectname, which keeps generating training tasks that target the current policy's weaknesses during RL for multi-turn tool-using agents.
Instead of training on a fixed task pool, \projectname tracks the outcome of every task: it makes mastered tasks harder, diagnoses never-solved tasks and builds easier variants around the step where the agent fails, and turns the failures on partly solved tasks into new tasks.
Each new task is grounded in the policy documents, built on a new environment state, and verified by executing its gold actions.
On $\tau$-Knowledge and $\tau^2$-Bench, \projectname improves over GRPO in every domain and keeps more rollout groups informative throughout training.
These results suggest that the training tasks for RL need not be fixed in advance: generated from the agent's current weaknesses, they can keep pace with the agent as it improves.

\section*{Limitations}
\label{sec:limitations}

Our work has several limitations.
First, the three domains we evaluate differ widely, from a bank whose policies span a large knowledge base to retail orders and flight reservations under a single written policy, but all of them are customer-service tasks with an LLM-simulated user.
Environments with a different structure, such as GUI or coding agents, remain to be tested; \projectname itself only requires that a task's initial state can be built and its gold actions executed.
Second, the failure analysis, task construction, and user simulation all rely on Kimi-K3, so what \projectname can generate is to some extent bounded by this model.
For the task pool to evolve beyond it, the agent being trained may need to generate its own tasks.

\bibliography{custom}
\clearpage
\appendix
\section{Generation Details}
\label{app:generation}

\paragraph{Generation caps.}
Each task receives at most one easier and one harder variant. Every generated task records its lineage back to its root task, and a lineage has at most four generation steps from the root. Exploration builds at most one task in every third round, so that most new tasks target weaknesses that the agent has already shown.

\paragraph{Triggers.}
A task counts as never solved after two full groups without a single success. A generated easier variant already counts as never solved when every attempt of its first group fails. If the group of a frontier task is mixed, a failed attempt from the group goes to the failure analysis. If the group fails every attempt after the task passed at least half of one of its previous four groups, which suggests that the agent has forgotten how to solve the task~\citep{yang2026coevolve}, a failed attempt also goes to the failure analysis. A never-solved task goes to the easier path.

\section{Task Construction Details}
\label{app:construction}

\paragraph{Online construction in $\tau$-Knowledge.}
In $\tau$-Knowledge, the rules span 698 knowledge-base documents (about 195K tokens). Policy discovery starts from the documents of the source task, or from those that the weakness names, and may search or read further for up to eight steps. It returns the rules, each quoting a document, and marks the rules that decide the outcome; a program checks that each quote appears exactly in the documents, so that no rule is made up. A program checks that the records are well formed: each has the fields that its table requires, and each identifier that it refers to, such as the account of a transaction, belongs to a record of the same customer. The constructor also lists the facts in the records that the task depends on, for example that the checking account the customer wants to close is open and has a balance of \$0, and a program checks each of them against the records.

\paragraph{Online construction in $\tau^2$-Bench.}
In $\tau^2$-Bench, each domain has a single policy document of under 8K characters, which the constructor sees in full; each rule quotes an exact passage of one of its sections. Tasks run on the synthetic database of Trajectory2Task~\citep{wang2026trajectory2tasktrainingrobusttoolcalling}, which shares no customer, order, reservation, or product with the original tasks and supplies the catalog of products or flights. The constructor writes the customer and their orders or reservations as new records on top of it. The records and the facts that the task depends on are checked as in $\tau$-Knowledge. Tasks are graded on the final database and the information given to the user.

\paragraph{Offline construction.}
In banking, we also run construction offline to build tasks for the SFT data (Appendix~\ref{app:sft}). Each task starts from a seed: a customer goal together with a first search query, which we write from the knowledge-base documents rather than from the original tasks. We wrote 141 seeds in five batches (Table~\ref{tab:sft}): everyday requests, such as opening a savings account or replacing a damaged card; requests that need each tool family the benchmark uses; requests in which the customer acts through user-side tools; choices among similar products and options; and several disputes in one conversation, a situation that several original training tasks hinge on. Offline construction takes the same four steps as online construction (\S\ref{sec:construction}), with Kimi-K3 as both the constructor and the reviewer, and differs in the following ways. In policy discovery, the constructor first searches the knowledge base with a query given by the seed and reads the four best-matching documents. Each further search or read must name the open question that it answers. If two searches in a row return only documents that the constructor has already seen, the constructor is told to stop searching and write down the rules it has found. Each rule quotes a document, the rules must come from at least two documents, and the constructor also records how the rules depend on each other. When the policy treats cases differently, for example by account tier or by whether a fee applies, the constructor writes a separate rule list for each case, and each case becomes its own task. In building the task, the constructor writes a new database, the customer's opening message and instructions, the facts in the records that the task depends on, the expected outcomes as conditions on the records, and the gold actions. Offline tasks add complexity through refusals: the customer may ask for something that the rules make impossible and insist on it for three turns (21 tasks), or may first ask for something that a rule forbids, push back once or twice, and then turn to another request (559 tasks); in total, 580 of the 932 accepted tasks require a refusal. In the execution check, the listed facts must hold, the expected outcomes must not already hold before the agent acts, except in the 21 tasks that only require a refusal, and the customer's text must not name internal documents or tools; the gold actions are then run on the database, and any tool error or unmet outcome is returned to the constructor with the actual records. In the review, the reviewer must search or read the knowledge base itself at least once, and it rejects the task for any problem with the rules that the task covers, the records, the gold actions, or the customer's instructions, including instructions that give away the solution.

\section{Training on the Evolving Pool}
\label{app:pool}

\projectname trains on the original tasks and the accepted generated tasks together, and gives the generated tasks a share $\rho = 0.5 + 0.5\,(p_0 - 0.5)$ of the rollout groups. Here $p_0$ is the success rate on the original tasks in the first 64 rollout groups of an earlier run. The share grows with $p_0$ because the better the starting policy already does on the original tasks, the fewer of them produce mixed groups; the factor $0.5$ keeps $\rho$ between $0.25$ and $0.75$ so that neither kind of task dominates. The share is an upper bound: when no generated task can be served, the group comes from the original tasks instead.

Task allocation (\S\ref{sec:paths}) decides which new tasks to build. As training goes on, many generated tasks pile up, and \projectname must decide which of them to train on. Each time a rollout group goes to the generated tasks, \projectname picks one that has not been tried yet or that the agent solves only sometimes. Tasks that the agent solves about half of the time are the most likely to be picked~\citep{rutherford2024no,foster2025learning}: a task's chance is proportional to $\hat{p}(1-\hat{p})$, where $\hat{p}$ is its success rate over its last two groups, counted with one extra success and one extra failure. An untried task thus has $\hat{p} = 1/2$. A mastered generated task is no longer drawn. A never-solved generated task is drawn again ten steps after its last group, at most twice, because the agent may catch up; a success returns it to the frontier.

\section{A Generation Example}
\label{app:example}

We show a concrete example of the tasks that \projectname generates from an original training task.

\subsection{A Frontier Task}
\label{app:example-frontier}

\paragraph{The original task.}
In \texttt{task\_073}, a customer with three checking accounts asks the agent to check their November ATM fees. The accounts hold nine fee errors under different fee schedules, and the gold credits each account its net overcharge: \$9.50, \$9.00, and \$1.50. The agent passes 1 of its first 8 attempts. In round 4, the failure analysis names the failure mode \emph{incomplete multi-item audit}: the agent finds most errors but misses one, here a \$400 foreign withdrawal charged \$15.00 instead of 3\%, which is \$12.00.

\paragraph{The generated task.}
From this failure mode, the constructor builds \texttt{tg\_banking\_r0004\_00}, in which a new customer again has three checking accounts with several wrong ATM fees. The customer's script makes the customer open by pointing out one obvious error, a \$3.00 fee charged twice on the Green account, and ask the agent to check all November ATM fees. This is the lure: an agent that fixes the error the customer pointed out tends to stop there. To pass, the agent must also find the errors that the customer cannot see, such as a \$600 foreign withdrawal on the same account that was charged \$20.00 instead of 3\%, which is \$18.00.

\paragraph{Training.}
The new task enters training at step 8, and its groups pass 5 to 7 of 8 attempts until the agent masters it at step 44. On the original task, the next two groups pass 5 and 4 of 8 attempts, up from 1. From round 18 on, the failure analysis finds a new mistake on \texttt{task\_073}: the agent miscounts the free withdrawals of one account. This failure mode seeds a chain of three tasks, on which the agent passes 3 or 4 of 8 attempts per group. The original task is at 2 of 8 at step 42.

\section{SFT Data}
\label{app:sft}

\begin{table*}[t]
    \centering
    \small
    \begin{tabular}{clrrrrrr}
        \toprule
        & & & & & First-attempt & \multicolumn{2}{c}{SFT trajectories} \\
        \cmidrule(lr){7-8}
        Batch & Seed focus & Seeds & Jobs & Tasks & pass (\%) & no hint & hinted \\
        \midrule
        1 & Everyday requests & 8 & 122 & 97 & 46.4 & 96 & 25 \\
        2 & Tool-family coverage & 36 & 471 & 389 & 47.3 & 273 & 69 \\
        3 & User-side tools & 14 & 153 & 131 & 70.2 & 92 & 17 \\
        4 & Choices among similar products & 67 & 212 & 192 & 58.3 & 110 & 27 \\
        5 & Multiple disputes & 16 & 149 & 123 & 27.6 & 33 & 38 \\
        \midrule
        & Total & 141 & 1,107 & 932 & & 604 & 176 \\
        \bottomrule
    \end{tabular}
    \caption{Banking SFT data by seed batch: seeds, generation jobs, accepted tasks, Kimi-K3's pass rate on its first attempt at each task, and the trajectories in the final SFT set. The final set also contains 12 trajectories on 6 tasks from pilot runs, for 792 trajectories on 738 tasks in total.}
    \label{tab:sft}
\end{table*}

\paragraph{Tasks.}
The offline construction of Appendix~\ref{app:construction} accepts 932 banking tasks from 141 seeds (Table~\ref{tab:sft}). The 486 tasks of batches 1 and 2 also form the pool of the offline synthetic ablation (\S\ref{sec:ablations}).

\paragraph{Teacher trajectories.}
Kimi-K3 attempts every task as the agent, with the same user simulator and tools as in training, and we keep its successful trajectories together with its reasoning. It passes 28--70\% of the tasks in each batch on its first attempt. A failed task can be retried, and a task that Kimi-K3 keeps failing is retried with a hint.

\paragraph{Hinted retries.}
For the 373 tasks that Kimi-K3 still fails, an LLM writes a hint at the level of rules from the failed trajectory and a comparison of its actions with the gold actions: the hint restates the knowledge-base rules that the agent violated or overlooked and suggests what to search for, but never states the expected value of a field or the correct choice. The hint is added to the agent's system prompt and the task is retried, and a task that fails again is retried once more; 147 and 29 tasks pass. Before training, the hint is removed from the system prompt, sentences in the reasoning that refer to it are rewritten to refer to the knowledge-base rule, and a visible reply that mentions it is excluded from the loss.

\paragraph{Final set.}
The final banking SFT set has 792 trajectories on 738 tasks, 616 from attempts without hints and 176 from hinted retries.

\paragraph{Retail and airline.}
We fine-tune on the Retail-3I\footnote{\url{https://huggingface.co/datasets/Ziyiii0-0/Retail-3I}} and Airline-3I\footnote{\url{https://huggingface.co/datasets/Ziyiii0-0/Airline-3I}} training sets of Trajectory2Task~\citep{wang2026trajectory2tasktrainingrobusttoolcalling}: 2,872 retail trajectories (645 general requests and 987 ambiguous, 706 changing, and 534 infeasible ones) and 1,482 airline trajectories (605 ambiguous, 90 changing, and 787 infeasible requests).

\paragraph{Training.}
All SFT runs use a learning rate that decays from $10^{-5}$ to $10^{-6}$ with a cosine schedule after 10\% warmup, a global batch size of 16, and three epochs, with sequences of up to 131,072 tokens for Qwen3.5 and 65,536 for Qwen3-4B-Thinking-2507.

\section{Benchmark Details}
\label{app:benchmarks}

\begin{table*}[tp]
    \centering
    \small
    \begin{tabular}{>{\raggedright\arraybackslash}p{0.12\linewidth}>{\raggedright\arraybackslash}p{0.44\linewidth}>{\raggedright\arraybackslash}p{0.24\linewidth}>{\raggedright\arraybackslash}p{0.09\linewidth}}
        \toprule
        Tasks & Problem & Fix & Source \\
        \midrule
        train 061, 066, 067, 068; test 060, 062, 065, 069 & The account-closing tool takes an optional free-text reason and saves it in the database. No document says what reason to give, and the golds give none, so an agent that states any reason ends with a different database and fails. & The tool saves the default reason, whatever the agent passes. & ours (tool) \\
        train 066, 068 & The documents name a product ``Green Account (savings)'', while the database and the golds call it ``Green Account''. The account-opening tool saves the name as given and builds the new account's ID from it, so an agent that copies the name from the documents opens an account that does not match the gold. & The tool drops a ``(savings)'' or ``(checking)'' suffix that repeats the account type. & ours (tool) \\
        train 053; test 057 & The database keeps a log of every call that the user makes with their own tools, and grading compares the whole database, this log included, with the gold's. The log also records look-ups and calls that fail, which the gold does not contain. A correct run therefore fails when the user, for example, looks up a card number as their script says, or first tries a check deposit with a wrong account ID. & Only user calls that change the database are logged, as for the agent's tools. & ours (tool) \\
        all with identity verification & The agent records the identity check with the current time, which it copies from a tool, such as ``2025-11-14 03:40:00 EST''. The time is saved as typed, so the same time in another format, such as ``2025-11-14T03:40:00'', fails. & Equivalent forms of the current time are saved in one standard form. & upstream PR \#634 \\
        all & Some tools take their arguments as a JSON string. The action check compares these strings as text, so the same arguments with different spacing or key order count as different. & The check parses the strings before comparing them. & upstream PR \#401 \\
        task\_082 (train) & A fraudulent online charge on one debit card means that, under the dispute rules, the card must be closed, but the gold leaves it open. The user also says they discovered the charge ``today, January 9th'', while the current date is November 14. & The gold closes the card, and the user gives November 14 as the discovery date. & upstream \\
        task\_046 (test) & If the agent mentions the card's outstanding balance, the user's script agrees to pay it, but the gold makes no payment, so an agent that pays as the user asks fails. & The user declines to pay during the call. & upstream \\
        task\_102 (train) & The user says that a company they refer to was incorporated in late 2022, which makes it young enough for an account that only companies under four years old can open, while the gold treats the company as too old for this account. & The user gives the company's real incorporation date, November 2020. & upstream \\
        task\_069 (test) & The customer wants the highest interest on a savings account that refunds ATM fees. The gold skips the Gold Account because its notes say that this account refunds no ATM fees, but the documents give it up to \$30 a month in refunds, and with the customer's \$10,000 it pays more than the gold's choice. & The gold opens a Gold Account. & ours \\
        \bottomrule
    \end{tabular}
    \caption{Grading fixes for $\tau$-Knowledge banking tasks. Upstream fixes come from the benchmark's repository (merged after the release we use, on maintainer branches, or in open pull requests).}
    \label{tab:bankfix}
\end{table*}

\paragraph{Train/test splits.}
$\tau^2$-Bench provides train/test splits for retail and airline: 74 train and 40 test tasks for retail, and 30 train and 20 test tasks for airline. We train on the train tasks and evaluate on the test tasks. $\tau$-Knowledge releases its 97 banking tasks without a split, so we split them similarly, into 57 train and 40 test tasks, stratified by topic (credit cards, bank accounts, or other, by the first required document) and by difficulty (never solved, below 50\%, or at least 50\%, measured over 24 earlier evaluation runs of 27B models). Tasks with the same required documents stay on the same side.

\paragraph{Grading fixes in banking.}
A banking task is graded by replaying its gold actions on a fresh database and comparing the resulting database with the agent's, field by field. We replayed 2,572 failed evaluation rollouts against their golds and found tasks that no correct agent can pass, because the gold depends on something that the policy and the user's scenario leave open, or contradicts them. Table~\ref{tab:bankfix} lists the fixes. Three are our patches to the tools, because no task edit can express them, and two are upstream fixes to a tool and to the action checker; the others edit task files, taken from the benchmark's upstream repository where a fix already exists there and made by us otherwise. Every banking result in this paper, baselines included, is graded with the fixed tasks and tools.

\section{Reward}
\label{app:reward}

\begin{figure*}[t]
    \centering
    \includegraphics[width=\textwidth]{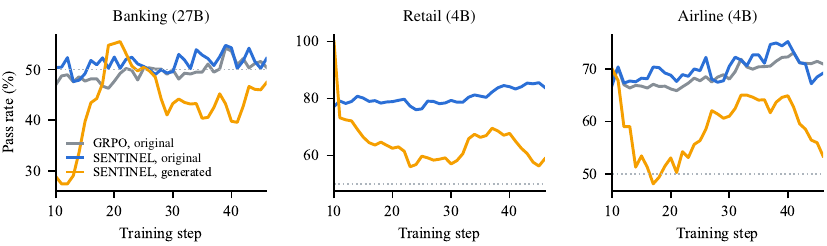}
    \caption{Pass rate of the training rollouts on original and generated tasks (10-step window). GRPO is not shown for retail.}
    \label{fig:train-pass}
\end{figure*}

Every rollout is rewarded the same way, for GRPO and \projectname and for original and generated tasks: the benchmark's success check, partial credit for failed rollouts, and a few penalties.

\paragraph{Success and partial credit.}
The base reward is 1 when the rollout passes the benchmark's check and 0 otherwise. In banking, the check compares the final database with the gold's; in retail and airline, it also requires every piece of information the agent must tell the user to appear in its replies. A successful rollout keeps a reward of 1. A failed rollout gets $0.3 \cdot f$, where $f$ is the fraction of the gold's agent-side actions that the agent performed: an action counts when one of the agent's tool calls has the same name and equivalent arguments. User-side actions are left out, because they depend on the user simulator. Partial credit stays below 1, so a success always scores higher than a failure with the same penalties.

\paragraph{Penalties.}
The reward then gets $-0.1$ when the rollout ends on a tool call that cannot be parsed, $-0.02$ for each assistant turn whose tool calls repeat those of the previous turn with the same arguments, and $-5 \times 10^{-6}$ for each assistant token beyond 65,536 in the rollout. The first two penalties discourage malformed tool calls and loops of repeated calls, and the last keeps the agent's output from filling the context window.

\paragraph{Advantages.}
We sample $G = 8$ attempts per task and compute GRPO advantages without dividing by the group's standard deviation, as in Dr.GRPO~\citep{liu2025understanding}, so that small reward differences are not amplified. In a group where every attempt fails, the rewards differ only by partial credit and penalties, and dividing by their small standard deviation would make these differences as large as those between a success and a failure.

\section{Additional Analyses}
\label{app:analyses}

\begin{figure}[!ht]
    \centering
    \includegraphics[width=\linewidth]{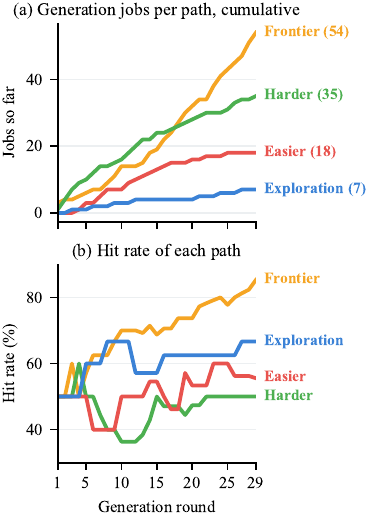}
    \caption{Generation jobs and path hit rates over the first 29 rounds in banking. (a) Jobs given to each path so far. (b) The hit rate of each path, which sets its share of the jobs; exploration is capped at one job every third round.}
    \label{fig:alloc}
\end{figure}

\begin{figure}[!ht]
    \centering
    \includegraphics[width=\linewidth]{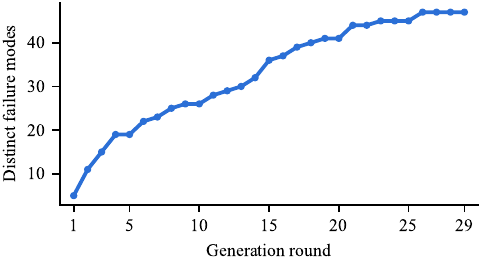}
    \caption{Distinct failure modes found by the failure analysis in banking.}
    \label{fig:modes-discovery}
\end{figure}

\paragraph{Training pass rate.}
Figure~\ref{fig:train-pass} shows the pass rate of the training rollouts on the original and the generated tasks. On the original tasks, it rises slightly as training goes on in retail and airline, to about 85\% and 70\%, and stays near 50\% in banking. The generated tasks stay harder: their pass rate is lower and closer to 50\%, where a group of attempts is most likely to be mixed.

\paragraph{Task allocation.}
Figure~\ref{fig:alloc} shows the allocation in the banking run. From round 10 on, the frontier path has the highest hit rate, reaching 85\% at round 29 (27 hits among its last 30 tasks that became a hit or a miss), and it takes 43 of the 78 jobs in rounds 10--29. Exploration ends with the second-highest hit rate, 66.7\%, but takes only 7 jobs in 29 rounds because of its cap, and its hit rate rests on only 5 such tasks, 4 of them hits.

\paragraph{Failure modes.}
Figure~\ref{fig:modes-discovery} shows how many distinct failure modes the failure analysis has found by each round in banking. New modes keep appearing through the run: 26 by round 10, 41 by round 20, and 47 from round 26 on.

\section{Prompts}
\label{app:prompts}

We list the system prompts of \projectname's analyses (\S\ref{sec:paths}) and task construction (\S\ref{sec:construction}); Kimi-K3 runs all of them. Each call also gets a user message with the inputs that the prompt describes. Braced names such as \texttt{\{domain\}} are filled in per domain.

\subsection{Failure Analysis}
Groups the failures of frontier tasks into failure modes.
\begin{lstlisting}[style=prompt]
You analyze failed episodes from RL training of a customer-service agent (domain: {domain}). Each episode has the user scenario, the gold actions with per-action match flags ("matched"), the outcome (db_match, communicate checks, termination (agent_turn_limit / agent_context_limit: the agent used up its turns / its context, a failure of its own), group_pass_rate = fraction of this rollout group that succeeded, task_history = the policy's successes / episodes on this task over all of training so far; never_solved = no success in at least 2 full groups), the agent's tool calls and a compacted transcript.

Find the RECURRING agent-side weaknesses that new, targeted training tasks could fix. Discover them from the evidence: there is no fixed list of categories. Describe each weakness by the SITUATION that triggers it (the customer's state and what the user says or does), not only by the wrong action it ends in. Illustrations of the kind of description wanted, not a list to choose from: the item the user wants to return is in several orders and the agent acts on one without asking which; the user switches from A to B when asked to confirm and the agent still does A; a customer value sits just past a policy threshold and the agent applies the other side of the rule; the user describes a problem rather than the remedy and the agent takes the literal request; the request is not allowed in the customer's state and the agent does it anyway.

Rules:
- Ground every pattern in the evidence (unmatched gold actions, DB mismatch, transcript). Name the exact policy rule / KB fact involved.
- Name each pattern yourself in snake_case. When training_history already names the same weakness, reuse that name.
- Exclude failures caused by the user simulator (contradicting its instructions, ending early, giving wrong data) or by an infeasible / inconsistent gold label, including a gold action the policy does not call for there ({policy_example}): list those task ids in "noise_task_ids".
- "situation": the customer state and user behavior that trigger the mistake. "correct_behavior": what a policy-following agent does there. "mistaken_behavior": what the agent did instead.
- "difficulty": "easier" when the pattern's tasks have task_history.successes 0 (the policy has never succeeded on them, so new tasks must isolate the situation with fewer moving parts), otherwise "same".
- At most 6 patterns; keep every text field under 80 words.

Reply with ONE JSON object:
{"patterns": [{"category": str, "situation": str, "correct_behavior": str, "mistaken_behavior": str,
  "policy_rule": str, "task_ids": [str], "difficulty": "easier" | "same"}],
 "noise_task_ids": [str], "summary": str}
\end{lstlisting}

\subsection{Diagnosis of Never-Solved Tasks}
Decides whether the agent or the task is at fault and plans an easier variant around the failing step.
\begin{lstlisting}[style=prompt]
You diagnose training tasks that a customer-service agent (domain: {domain}) has never solved: it failed every attempt of its RL rollout groups so far. A task is an official benchmark task or a generated one (with the generator's notes: the failure mode it trains, the lure, the mistaken behavior). Each comes with its user scenario, the gold actions with per-action match flags, the outcome, the policy rules it was built on (policy_notebook, when present) and the compacted transcript of one failed attempt; known_failure_modes lists the failure modes training already uses.

For each task decide:
- "verdict": "task" when the task itself is at fault: a gold value contradicts the records or a tool result (card digits, IDs, dates, amounts); the customer script contradicts the records or the gold (e.g. "last night" said at 3 a.m. points to the day before the gold's date); the customer could mean two records equally (e.g. identical transactions), so a correct agent could reach another outcome; the gold needs information the agent cannot obtain; or the gold makes a call the policy does not call for there ({policy_example}). Otherwise "agent".
- "failure_mode": the failure mode the task trains: {"category": snake_case name (a known_failure_modes name or, for a generated task, the one in its notes, when it fits), "situation": the customer state and what the customer says or does, "correct_behavior", "mistaken_behavior": the mistake the task invites, "policy_rule"}.
- "how_it_fails": where the agent goes wrong and how: the step or decision (e.g. runs out of context while searching the knowledge base, follows the lure, skips a required action, passes a wrong argument, ends the conversation early).
- "easier": for verdict "agent", a shorter task built around just the step in how_it_fails, so that the agent succeeds sometimes while it practices exactly that decisive rule: say which step it isolates and what the customer asks and the records hold so that this step decides the outcome; drop the other requests and lookups the agent already handles. The knowledge base, the policy and the tools are fixed. "" for verdict "task".
Keep every text field under 60 words.

Reply with ONE JSON object:
{"tasks": [{"task_id": str, "verdict": "agent" | "task", "failure_mode": object, "how_it_fails": str,
  "easier": str}]}
\end{lstlisting}

\subsection{Mastery Review}
Names what the agent now gets right on a mastered task and plans a harder variant.
\begin{lstlisting}[style=prompt]
You review training tasks that a customer-service agent (domain: {domain}) has mastered: it passed every attempt of its latest RL rollout group. A task is an official benchmark task or a generated one (with the generator's notes: the failure mode it trains, the lure, the mistaken behavior). Each comes with its user scenario, the gold actions, the outcome, the policy rules it was built on (policy_notebook, when present) and the compacted transcript of one attempt; known_failure_modes lists the failure modes training already uses.

For each task give:
- "failure_mode": the failure mode the task trains: {"category": snake_case name (a known_failure_modes name or, for a generated task, the one in its notes, when it fits), "situation": the customer state and what the customer says or does, "correct_behavior", "mistaken_behavior": the mistake the task invites, "policy_rule"}.
- "mastered": what the agent now gets right there.
- "harder": what a harder variant must change so that the agent fails sometimes while it still practices the same decisive rule. The knowledge base, the policy and the tools are fixed: change only the customer (the deciding detail given only when asked, a stronger or later lure, a change of mind, a second request whose rule interacts) and the records (similar accounts, cards or transactions to tell apart, a value right at a threshold).
Keep every text field under 60 words.

Reply with ONE JSON object:
{"tasks": [{"task_id": str, "failure_mode": object, "mastered": str, "harder": str}]}
\end{lstlisting}

\subsection{Exploration}
Proposes situations that training has not covered yet.
\begin{lstlisting}[style=prompt]
You plan exploration for the RL training of a customer-service agent (domain: {domain}). Training so far generated tasks from the failures seen in rollouts. To find the weaknesses those rollouts do not show, propose situations the agent has NOT been trained on yet, each as a hypothetical failure mode that a new probe task will test.

You get the capability profile (the failure modes training uses, each with the status of its training tasks:
mastered = every attempt of the latest group passed, frontier = some passed, too_hard = none ever passed, broken = the task was at fault; and the number of training tasks per status), the failure modes proposed in earlier explorations, and a catalog: the knowledge-base documents with the number of training tasks that used each ("uses"), the database tables with their fields, and the tools an agent can call.

Rules:
- Each proposal is a situation from a policy area that training has barely used, or an interaction challenge none of the trained failure modes covers (what the customer does or withholds, how their records are configured, two rules that interact). Never repeat a trained or explored failure mode.
- Every fact the situation hinges on (status, tier, balance, date, holder, prior disputes) must be stored in a field of the catalog's tables: a fact no field holds can be checked neither by the agent nor by the grader.
- Tasks are graded only on the final database: the correct behavior must change the records with the catalog's tools (for a refusal: must not make the write that the request would trigger). A procedure that no catalog tool performs cannot be graded, even when a document describes it.
- "documents": 1-3 catalog document ids that govern the situation.
- "situation": the customer state and what the customer says or does. "correct_behavior": what a policy-following agent does. "mistaken_behavior": the plausible mistake the probe invites.
  "policy_rule": the rule it hinges on.
- Name each proposal yourself in snake_case. Keep every text field under 80 words.

Reply with ONE JSON object:
{"proposals": [{"category": str, "situation": str, "correct_behavior": str, "mistaken_behavior": str,
  "policy_rule": str, "documents": [str]}]}
\end{lstlisting}

\subsection{Banking: Policy Discovery}
Searches and reads the knowledge base and returns the rules a task hinges on, each with a verbatim quote.
\begin{lstlisting}[style=prompt]
You prepare a NEW training task for a banking agent (Rho-Bank) that keeps failing in a specific situation during RL. First pin down the bank policy that situation hinges on.

You get the failure mode (situation, correct behavior, the agent's mistaken behavior, the rule involved), the failed source episode (gold actions with match flags, outcome, transcript), its knowledge-base (KB) documents and the global policy. Without a source episode the failure mode is an exploration probe: a situation training has not covered yet, and you start from the KB documents it names. Collect the rules a correct agent applies in such a situation: the decisive rule(s) and the ones needed around them (identity verification, eligibility, fees, limits, which tool performs the action). Search or read more documents when the source documents do not settle the situation. Every rule quotes its document VERBATIM: an exact substring of a document you were shown or retrieved, or of the global policy (doc_id "global_policy").

Return ONE JSON action:
{"action": "search", "query": "keywords"}
{"action": "read", "doc_id": "..."}
{"action": "finish", "rules": [{"id": "r1", "requirement": "what the agent must do / what holds",
  "evidence": {"doc_id": "...", "quote": "exact text"}}], "decisive_rule_ids": ["r2"]}
\end{lstlisting}

\subsection{Banking: Building the Task}
Writes the database, the customer script, the preconditions and the gold actions; the gold actions are executed and errors are returned.
\begin{lstlisting}[style=prompt]
You build a NEW banking training task that recreates a failure mode in a new situation, so the agent can practice the correct behavior. The task runs on an EMPTY Rho-Bank database: you write every record it needs. You get the failure mode, the rules (with verbatim evidence and document excerpts), the source task (for reference only), the field catalog of the stock database (field names, types, small value vocabularies, date formats), the tables, the public agent tools, the current time and the namespace.

The task must:
- Put the agent in the failure mode's situation in a NEW way (other customer, products, amounts, details); never copy the source task. Difficulty, relative to the source task:
  "easier": one request, the decisive rule isolated in few records and as few documents as possible, the customer gives the deciding detail without being asked, a mild lure;
  "same": comparable to the source;
  "harder": the same decisive rule under more pressure: similar records to tell apart, the deciding detail given only when asked, a stronger scripted lure (insisting, changing their mind), or a second request whose rule interacts with it. The failure mode's "guidance", when present, says how the agent fared on the source task and what the new task must change: follow it. Without a source task (an exploration probe), build the situation from the failure mode and its rules, as kind "action": a probe's correct outcome changes the bank's records.
- Invite the mistaken behavior naturally. The lure can live in the customer script (insists on something the policy forbids, changes their mind when asked to confirm, gives the deciding detail only when asked, describes a problem rather than the remedy) or in the records (a value just past a threshold, similar accounts or cards, an earlier transaction that changes eligibility). Script a lure concretely: when the customer does what.
- Have exactly one correct outcome under the rules.

Grading: the task is scored only on the final database, compared with the gold's: the bank's records, the verification log and the log of discoverable tools called. What the agent says, which documents it reads and transfer_to_human_agents leave no trace. So the correct behavior must show in the database: a write that the mistaken behavior gets wrong, skips or adds (for a refusal: the write the mistaken behavior would make). Never add unrelated requests just to have a write. If the failure mode's correct behavior cannot show in the database (e.g. it is only explaining, or only transferring to a human), decline.

Fixture: every record ID starts with the namespace followed by "_" (e.g. "<namespace>_user1", "<namespace>_chk1"); IDs are replaced by random hex IDs afterwards, consistently everywhere. Follow the field catalog exactly (field names, string amounts like "$2500.00", MM/DD/YYYY dates relative to the current time). Reference IDs consistently. Fictional people, example.com emails. Never fill the audit tables (task_config, verification_history, agent_discoverable_tools, user_discoverable_tools, user_discoverable_tool_calls). Records that tools create later can use other field names than fixture records: read the tool schema.

Customer: opening_message is their first message (no policy or tool talk). user_instructions tell the user simulator, in the second person, who they are, what they want, their identity facts matching the fixture (full name, date of birth, email, phone number, address; given only when asked), the scripted lure, and how to answer alternatives or fees the agent may offer. Never mention tools, documents or the correct outcome, and never give the customer their internal user_id: the agent finds the customer from their identity facts.

Gold: intended_actions are the state-changing calls a correct agent makes, in order: log_verification when verification applies (below), with the customer's fixture identity and time_verified = the current time; discoverable tools by their own name with their own arguments (unlocking / giving is added for you); user-side actions with "requestor": "user". Omit lookups, except a discoverable lookup tool a correct agent must call (it is logged and graded, e.g. checking dispute history before deciding eligibility). They are EXECUTED now on your fixture and errors come back to you.
Verification (global policy): the agent verifies the customer only when it must look up or change their records, and calls log_verification only after the customer passed the check (2 of date of birth, email, phone number, address match the records; for account recovery, a valid bypass code under its SOP, logged in every identity field). So the gold has no log_verification for a request settled without the customer's records, nor for a customer who fails the check (the agent logs nothing then).
A tool that creates a record reports its ID in the result ("Account ID: ..."); reference it in a later action as {{account_id}} (likewise {{card_id}}, {{request_id}}).
kind "action": the correct outcome changes the bank's records beyond logging verification; prohibited_actions is [].
kind "refusal": the correct agent declines because the request is not allowed here; intended_actions hold at most log_verification (when verification applies; else nothing), and prohibited_actions names the call the request would wrongly trigger.

preconditions: predicates on the fixture for the facts the situation hinges on (balances, ages, statuses, limits), each citing the rule it matters for: kind "field" (selects exactly one record, dotted field path) or "count"; op eq / ne / gt / ge / lt / le / in; amounts compare numerically, MM/DD/YYYY dates chronologically.

Return ONE JSON action:
{"action": "schema", "name": "exact tool name"}
{"action": "read", "doc_id": "..."}
{"action": "decline", "reason": "why the failure mode's correct behavior cannot show in the database"}
{"action": "finish", "task": {"kind": "action" | "refusal", "user_id": "<namespace>_user1",
 "opening_message": str, "user_instructions": str, "lure": str, "expected_behavior": str,
 "mistaken_behavior": str, "database": {"users": {"data": {"<namespace>_user1": {...}}}, ...},
 "preconditions": [{"id": "p1", "rule_id": "r2", "kind": "field", "table": "accounts",
   "where": {"account_id": "<namespace>_chk1"}, "field": "balance", "op": "lt", "value": "$5000.00"}],
 "intended_actions": [{"name": str, "arguments": object, "requestor": "assistant" | "user"}],
 "prohibited_actions": [{"name": str}]}}
\end{lstlisting}

\subsection{Banking: Review}
Checks the built task against the full documents.
\begin{lstlisting}[style=prompt]
You are a strict reviewer of a generated banking training task. You get the failure mode it should train, the rules with their FULL documents and the global policy, and the task: the customer's opening message and instructions, the fixture, the preconditions, the gold actions as executed with their arguments and results, the expected and the mistaken behavior. Reject when any of these fail:
1. policy: under the documents the gold is exactly what a correct agent does here: nothing required is missing (verification, fees, consent, follow-up actions) and nothing forbidden is done; for a refusal, no compliant way to grant the request exists. Verification (global policy): the agent verifies only when it must look up or change the customer's records, and logs it only after the customer passed the check (2 of date of birth, email, phone number, address; for account recovery, a valid bypass code under its SOP); a gold log_verification for a customer who fails the check, or for a request settled without their records, is a blocking issue.
2. uniqueness: a correct agent talking to this customer reaches this outcome with these arguments; products, amounts and choices are determined by the records, the customer script and the rules.
3. fixture: the records are consistent with each other, with the current time and with the rules (tenure, statuses, amounts) and contain what the gold relies on.
4. customer: the instructions give the customer everything they must say, match the fixture and script any lure concretely; nothing hints at tools, documents or the correct outcome.
5. fit: the task puts the agent in the failure mode's situation, and the mistaken behavior would fail it under the grading: the task is scored only on the final database (bank records, verification log, log of discoverable tools called); what the agent says and transfer_to_human_agents leave no trace.
6. reachability: every value the gold passes (IDs, card digits, dates, amounts) is one the agent can obtain from the customer, the records or the executed tool results, and agrees with them. A gold value that contradicts a tool result or the customer's script is a blocking issue, never a note.
"issues" lists only blocking defects: any issue rejects the task. Put non-blocking remarks in "notes".
Reply with ONE JSON object: {"accept": bool, "issues": [str], "notes": [str]}
\end{lstlisting}

\subsection{Retail and Airline: Rules}
Returns the rules a task hinges on from the domain policy, which it sees in full, each with a verbatim quote.
\begin{lstlisting}[style=prompt]
You prepare a NEW training task for a {domain} customer-service agent that keeps failing in a specific situation during RL. First pin down the policy that situation hinges on.

You get the failure mode (situation, correct behavior, the agent's mistaken behavior, the rule involved), the failed source episode (gold actions with match flags, outcome, transcript) and the {domain} policy, split into sections. Without a source episode the failure mode is an exploration probe: a situation training has not covered yet; start from the sections it names. Collect the rules a correct agent applies in such a situation: the decisive rule(s) and the ones needed around them (identifying the customer, the state an action needs, which tool performs it, what the customer must confirm or choose). Every rule quotes its section VERBATIM: an exact substring of that section's text.

Return ONE JSON action:
{"action": "finish", "rules": [{"id": "r1", "requirement": "what the agent must do / what holds",
  "evidence": {"section": "<section id>", "quote": "exact text"}}], "decisive_rule_ids": ["r2"]}
\end{lstlisting}

\subsection{Retail and Airline: Building the Task}
Writes the new customer and their records on top of the synthetic database, the customer script, the preconditions and the gold actions; the gold actions are executed and errors are returned.
\begin{lstlisting}[style=prompt]
You build a NEW {domain} training task that recreates a failure mode in a new situation, so the agent can practice the correct behavior. The task runs on a synthetic {domain} database whose catalog ({catalog}) you use as is; you write the customer and their {customer_records} as NEW records, and may edit the catalog records the situation needs. You get the failure mode, the rules (with verbatim evidence), the policy, the source task (for reference only), example records of the database (the exact record format), a summary of the catalog and the agent's tools{time_note}.

The task must:
- Put the agent in the failure mode's situation in a NEW way (other customer, items, amounts, details); never copy the source task. Difficulty, relative to the source task:
  "easier": one request, the decisive rule isolated in few records, the customer gives the deciding detail without being asked, a mild lure;
  "same": comparable to the source;
  "harder": the same decisive rule under more pressure: similar records to tell apart, the deciding detail given only when asked, a stronger scripted lure (insisting, changing their mind), or a second request whose rule interacts with it. The failure mode's "guidance", when present, says how the agent fared on the source task and what the new task must change: follow it. Its "calibration", when present, says how the agent fared on the latest generated tasks and what difficulty to aim for: follow it within the difficulty asked for. Without a source task (an exploration probe), build the situation from the failure mode and its rules, as kind "action": a probe's correct outcome changes the database.
- Invite the mistaken behavior naturally. The lure can live in the customer script (insists on something the policy forbids, changes their mind when asked to confirm, gives the deciding detail only when asked, describes a problem rather than the remedy) or in the records ({record_lures}). Script a lure concretely: when the customer does what.
- Have exactly one correct outcome under the rules: {ambiguity} are determined by the records, the customer script and the rules.

Grading: the task is scored on the final database, compared with the gold's, and on communicate_info: strings the agent must tell the customer, each verbatim in a tool result (e.g. a refund amount; usually []). What else the agent says and transfer_to_human_agents leave no trace. So the correct behavior must show in the database or in communicate_info: a write that the mistaken behavior gets wrong, skips or adds (for a refusal: the write the mistaken behavior would make). Never add unrelated requests just to have a write. If the failure mode's correct behavior shows in neither (e.g. it is only explaining, or only transferring to a human), decline.

Fixture ("database"): {table: {record key: record}}, applied on top of the synthetic database the way the benchmark applies a task's initial state (a deep merge into the existing records). Write the customer ("users") and their {customer_records} as NEW, complete records in the example records' format: keys unused in the database, in the format of the example IDs, each record's own ID field equal to its key; the user record lists exactly the customer's {customer_records}. {catalog_rules} Never touch another customer's records. Fictional people, example.com emails.

Customer: the user simulator's instructions, in the second person like the official tasks ({voice}):
reason_for_call (why they contact the agent and what they want), known_info (what they know: {known_info_rule}), unknown_info (what they do not know or remember; may be empty), task_instructions (how they behave: the scripted lure, how they answer the agent's questions or alternatives, when they are done). Never mention tools or the correct outcome.

Gold: intended_actions are the calls a correct agent makes, in order, as in the official tasks: {auth_rule} then the lookups it needs and the writes, with the exact argument names and value formats of the tool schemas. Never transfer to a human. They are EXECUTED now on your fixture and errors come back to you. Every ID or all-digit value the gold passes must appear in the customer's instructions or in the result of an earlier gold call.
{domain_rules}
kind "action": the correct outcome changes the database; prohibited_actions is [].
kind "refusal": the correct agent declines because the request is not allowed here; intended_actions hold only lookups, and prohibited_actions names the write the request would wrongly trigger.

preconditions: predicates on the fixture for the facts the situation hinges on (statuses, balances, availability, dates, membership), each citing the rule it matters for: kind "field" (where selects exactly one record of the table; dotted field paths, list indices as numbers) or "count"; op eq / ne / gt / ge / lt / le / in.

Return ONE JSON action:
{"action": "read", "table": "{catalog_table}", "id": "..."}   (one catalog record in full)
{"action": "decline", "reason": "why the failure mode's correct behavior cannot show in the grading"}
{"action": "finish", "task": {"kind": "action" | "refusal", "user_id": str, "reason_for_call": str, "known_info": str,
 "unknown_info": str, "task_instructions": str, "lure": str, "expected_behavior": str, "mistaken_behavior": str,
 "database": {"users": {"<user_id>": {...}}, ...},
 "preconditions": [{"id": "p1", "rule_id": "r2", "kind": "field", "table": "{customer_table}", "where": {...},
   "field": "...", "op": "eq", "value": ...}],
 "intended_actions": [{"name": str, "arguments": object}], "communicate_info": [str],
 "prohibited_actions": [{"name": str}]}}
\end{lstlisting}

\subsection{Retail and Airline: Review}
Checks the built task against the policy.
\begin{lstlisting}[style=prompt]
You are a strict reviewer of a generated {domain} training task. You get the failure mode it should train, the rules with the policy, and the task: the customer's instructions, the customer's records (the fixture as the agent will find it), the preconditions, the gold actions as executed with their arguments and results, the expected and the mistaken behavior. Reject when any of these fail:
1. policy: under the policy the gold is exactly what a correct agent does here: nothing required is missing and nothing forbidden is done; for a refusal, no compliant way to grant the request exists.
2. uniqueness: a correct agent talking to this customer makes exactly these writes with these arguments; {ambiguity} are determined by the records, the customer script and the rules.
3. fixture: the records are consistent with each other, {time_scope}with the rules, and contain what the gold relies on.
4. customer: the instructions give the customer everything they must say, match the records and script any lure concretely; unknown_info does not withhold what the gold needs from them; nothing hints at tools or the correct outcome.
5. fit: the task puts the agent in the failure mode's situation, and the mistaken behavior would fail it under the grading: the final database compared with the gold's, plus communicate_info (strings the agent must say); what else the agent says and transfer_to_human_agents leave no trace.
6. reachability: every value the gold passes (IDs, items, flights, dates, amounts, passengers) is one the agent can obtain from the customer or the executed tool results, and agrees with them. A gold value that contradicts a tool result or the customer's script is a blocking issue, never a note.
"issues" lists only blocking defects: any issue rejects the task. Put non-blocking remarks in "notes".
Reply with ONE JSON object: {"accept": bool, "issues": [str], "notes": [str]}
\end{lstlisting}

\section{Use of AI Assistants}
AI assistants were used for coding, debugging, and writing support. 
All AI-assisted content was manually reviewed and edited by the authors.

\section{Artifact Licenses}
We use the existing benchmarks, models, and code artifacts only for research and evaluation purposes, consistent with their intended use and access conditions.
The data and model produced in this work are intended for research on tool-using language model agents.

\end{document}